\documentclass[letterpaper, 10 pt, conference]{ieeeconf}  

\IEEEoverridecommandlockouts                              

\usepackage[pdftex]{graphicx}
\usepackage{multirow}
\usepackage{marvosym}
\usepackage{amssymb}
\usepackage{url}
\usepackage{cite}
\usepackage{makecell}
\usepackage{subfigure}

\graphicspath{{Figure/}}

\title{\LARGE \bf
Reveal of Domain Effect: How Visual Restoration Contributes to \\ Object Detection in Aquatic Scenes
}

\author{Xingyu Chen, Yue Lu, Zhengxing Wu, Junzhi Yu, and Li Wen
\thanks{This work was supported by the National Natural Science Foundation of China (nos. 61633004, 61633020, 61973303, 61633017, and 61725305).}
\thanks{X. Chen, Z. Wu, Y. Lu, and J. Yu are with the State Key Laboratory of Management and Control for Complex Systems, Institute of Automation, Chinese Academy of Sciences, Beijing 100190, China and the School of Artificial Intelligence, University of Chinese Academy of Sciences, Beijing 100049, China. J. Yu is also with the State Key Laboratory for Turbulence and Complex Systems, Department of Mechanics and Engineering Science, BIC-ESAT, College of Engineering, Peking University, Beijing 100871, China
        {\tt\small \{chenxingyu2015, luyue2018, zhengxing.wu, junzhi.yu\}@ia.ac.cn}}%
\thanks{L. Wen is with the School of Mechanical Engineering and Automation, Beihang University, Beijing 100191, China,
        {\tt\small liwen@buaa.edu.cn}}
}
\begin{document}
\maketitle
\thispagestyle{empty}
\pagestyle{empty}

\begin{abstract}
Underwater robotic perception usually requires visual restoration and object detection, both of which have been studied for many years. Meanwhile, data domain has a huge impact on modern data-driven leaning process. However, exactly indicating domain effect, the relation between restoration and detection remains unclear. In this paper, we generally investigate the relation of quality-diverse data domain to detection performance. In the meantime, we unveil how visual restoration contributes to object detection in real-world underwater scenes. According to our analysis, five key discoveries are reported: 1) Domain quality has an ignorable effect on within-domain convolutional representation and detection accuracy; 2) low-quality domain leads to higher generalization ability in cross-domain detection; 3) low-quality domain can hardly be well learned in a domain-mixed learning process; 4) degrading recall efficiency, restoration cannot improve within-domain detection accuracy; 5) visual restoration is beneficial to detection in the wild by reducing the domain shift between training data and real-world scenes. Finally, as an illustrative example, we successfully perform underwater object detection with an aquatic robot.
\end{abstract}

\section{Introduction}
{\noindent\bf Background.} Within the last few years, great efforts have been made for underwater robotics. For example, Gong \emph{et~al.} designed a soft robotic arm for underwater operation \cite{bib:SoftArm}. Cai \emph{et~al.} developed a hybrid-driven underwater vehicle-manipulator for collecting marine products \cite{bib:Manipulator}. Towards intelligent autonomous robots, visual methods are usually adopted for underwater scene perception \cite{bib:SoftArm,bib:Manipulator,bib:ImageServoing,bib:VisualSLAM}.

With the advent of convolutional neural network (CNN), object detection has been a surging topic in computer vision \cite{bib:SSD,bib:RetinaNet,bib:RefineDet,bib:DRN}, and in the meantime, object detection is a fundamental tactic for robotic perception \cite{bib:TDR}. Based on detection, robots can discover what and where the target is. However, because of optical absorption and scattering, underwater visual signal usually suffers from degeneration and forms low-quality images/videos \cite{bib:Sc04}. Note that low quality means low contrast, high color distortion, and strong haziness. Therefore, visual restoration has been widely studied \cite{bib:Sc04,bib:UnderwaterBench,bib:Li16,bib:Pe17,bib:Ch17,bib:GAN-RS} so that visual quality can be improved for subsequent image processing. By and large, visual restoration and object detection are two essential abilities for an aquatic robot to perform object perception.

\begin{figure}[!t]
\centering
\includegraphics[width=8.cm]{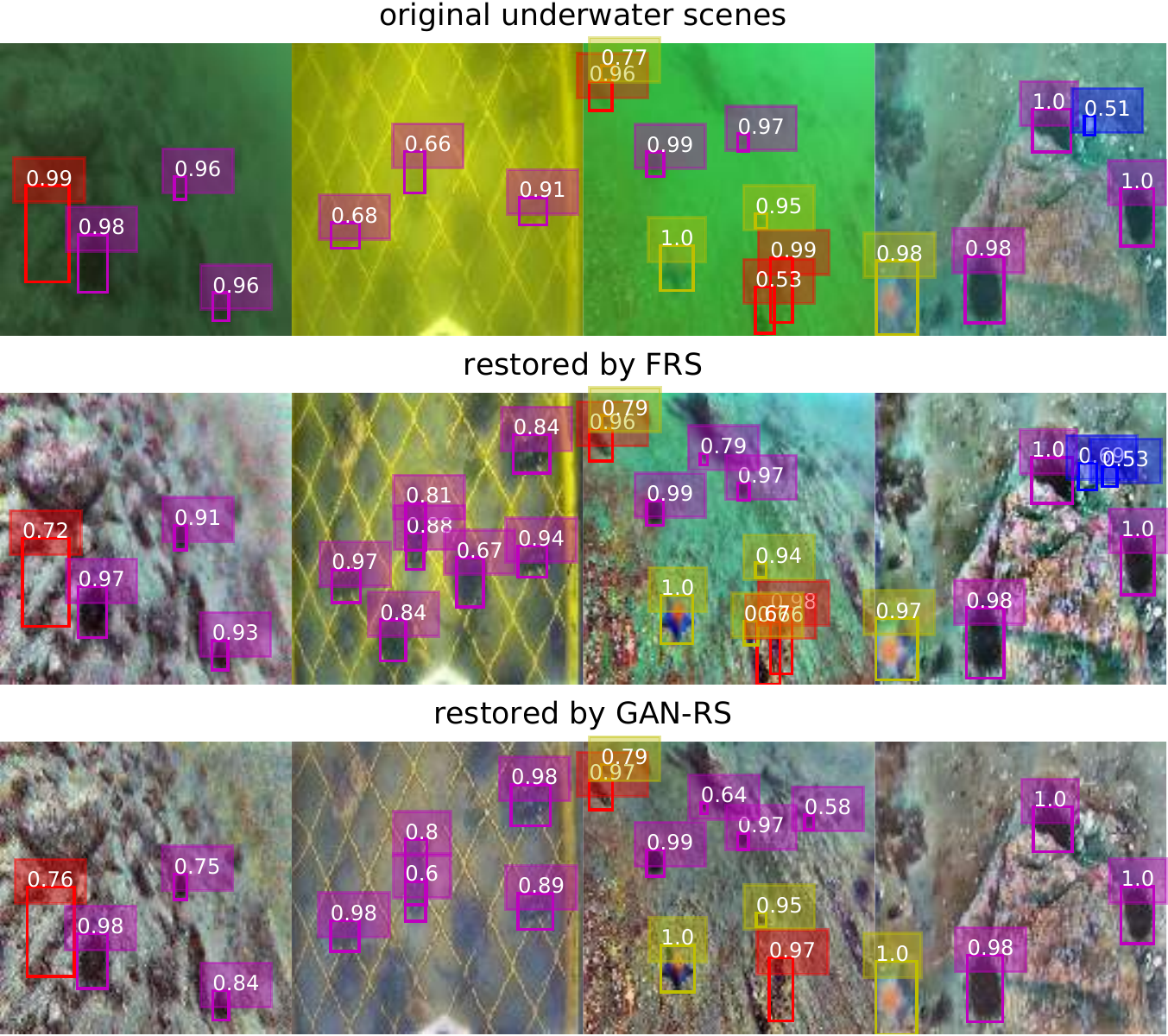}
\caption{Underwater object detection based on different restoration manners. Underwater visual degeneration is diverse as shown in the top line. Relieving this degeneration, FRS \cite{bib:Ch17} and GAN-RS \cite{bib:GAN-RS} generate clear images. Further, for the same scenario, different detection results are produced because of different restoration methods. Colors differentiate categories.}
\label{fig:int}
\end{figure}

{\noindent\bf Problem \& motivation.} Although visual restoration has proven to be helpful for traditional man-made features (e.g., SIFT \cite{bib:SIFT}) \cite{bib:Li16}, the relation between image quality and convolutional representation remains unclear. As demonstrated in Fig.~\ref{fig:int}, underwater scenes are always degenerated, and moreover, the degeneration usually has different styles, i.e., color distortion, haziness, and illumination (see the top line). By filtering-based restoration (FRS) \cite{bib:Ch17} and GAN-based restoration (GAN-RS) \cite{bib:GAN-RS}, higher-quality images are generated. Although each column of Fig.~\ref{fig:int} is the same scenario, their detection results are diverse with DRN detector \cite{bib:DRN}. Therefore, scopes of restoration and detection should have latent relevance that should be investigated. To this end, we study to answer a question $-$ \emph{how does visual restoration contribute to object detection in aquatic scenes?}

In addition, visual restoration exactly produces the change of data domain, and it is known that data domain is important for data-driven learning process \cite{bib:DomainRCNN,bib:DomainDet,bib:SubAlign,bib:DomainAdapt,bib:CrossWeakly}. However, under the condition of different data domains, within-domain and cross-domain detection performances have rarely been studied. That is, the domain effect on object detection remains unclear. In our opinion, exploring the effect of data domain is instructive for building robust real-world detectors. Thereby, we are motivated to investigate the relation between image quality and detection performance based on visual restoration to unveil domain effect on object detection. In this way, the relation of restoration to detection can also be exposed.


{\noindent\bf Our work.} In this paper, we joint analyze visual restoration and object detection for underwater robotic perception. At first, we construct quality-diverse data domains with FRS and GAN-RS for both training and testing. FRS is a traditional filtering method and GAN-RS is a learning-based scheme, so they can be representative for the restoration sphere. Further, we investigate typical single-stage detectors (i.e., SSD \cite{bib:SSD}, RetinaNet \cite{bib:RetinaNet}, RefineDet \cite{bib:RefineDet}, and DRN \cite{bib:DRN}) on different data domains, then within-domain and cross-domain performances are analyzed. Finally, real-world experiments are conducted on the seabed for online object detection. Based on our study, the relation of restoration-based data domain to detection performance is unveiled. As a result, although it induces adverse effects on object detection, visual restoration efficiently suppresses domain shift (i.e., discordance between training domain and testing domain) between training images and practical scenes. Thus, visual restoration still plays an essential role in aquatic robotic perception. Our contributions are summarized as follows:
\begin{itemize}
  \item We reveal three domain effects on detection: 1) Domain quality has a negligible effect on within-domain convolutional representation and detection accuracy after sufficient training; 2) low-quality domain brings about better generalization in cross-domain detection; 3) in domain-mixed training, low-quality domain can hardly be well learned.
  \item We indicate that restoration is a thankless operation for improving within-domain detection accuracy. In detail, it reduces recall efficiency \cite{bib:SelectRefine}. However, visual restoration is beneficial in reducing domain shift between training data and practical aquatic scenes so that online detection performance can be boosted. Therefore, it is an essential operation in real-world object perception.
  \item Based on our analysis, online object detection is successfully conducted on the field unstructured seabed with an aquatic vision-based robot.
\end{itemize}

\section{Related Work}
\label{sec:RW}
{\noindent\bf Underwater visual restoration.} Because of natural physical phenomenon, underwater visual signal is usually degenerated, forming low-quality vision. In detail, underwater image/video shows low contrast, high color distortion, and strong haziness, making image processing difficult. Schechner and Karpel attributed this degeneration to visual absorption and scattering \cite{bib:Sc04}. Overcoming this difficulty, Peng and Cosman proposed a restoration method based on image blurriness and light absorption, which estimated scene depth for image formation model \cite{bib:Pe17}. Chen \emph{et~al.} adopted filtering model and artificial fish algorithm for real-time visual restoration \cite{bib:Ch17}. Li \emph{et~al.} hierarchically estimated background light and transmission map, and their method was characterized by minimum information loss \cite{bib:Li16}. Chen~\emph{et~al.} proposed a weakly supervised GAN and an adversarial critic training to achieve real-time adaptive restoration \cite{bib:GAN-RS}. Recently, Liu~\emph{et~al.} built an underwater enhancement benchmark for follow-up works, whose samples were collected on the seabed under natural light \cite{bib:UnderwaterBench}.

With the above-mentioned studies, it is revealed that visual restoration is beneficial in clearing image details and producing salient low-level features. For example, canonical SIFT \cite{bib:SIFT} algorithms deliver a huge performance improvement based on restoration \cite{bib:Li16}. However, how visual restoration contributes to CNN-based feature representation remains unclear. Moreover, visual restoration is tightly related to data domain, so we explore domain effect based on restoration.

{\noindent\bf Object detection \& domain adaption.} During the deep learning era, single-stage object detection uses a single-shot network for regression and classification. As a pioneering work, Liu \emph{et~al.} proposed SSD for real-time detection\cite{bib:SSD}. Inspired by feature pyramid network, Li \emph{et~al.} developed RetinaNet to propagate CNN features in a top-down manner for enlarging shallow layers' receptive field \cite{bib:RetinaNet}. Zhang \emph{et~al.} introduced two-step regression to the single-stage pipeline and designed RefineDet for addressing class imbalance problem. Chen \emph{et~al.} proposed DRN with anchor-offset detection that achieved single-stage region proposal \cite{bib:DRN}. Although some two-stage detectors \cite{bib:ACoupleNet} and anchor-free detectors \cite{bib:ExtremeNet} could induce higher accuracy, the single-stage methods maintain a better accuracy-speed trade-off for robotic~tasks.

Above detectors generally assume that training and test samples fall within an identical distribution. However, real-world data usually suffer from domain shift, which affects detection performance. Hence, cross-domain robustness of object detection is recently explored. Chen \emph{et~al.} proposed adaptive components for image-level and instance-level domain shift based on $\mathcal{H}$-divergence theory \cite{bib:DomainRCNN}. Xu \emph{et~al.} utilized deformable part-based model and adaptive SVM for mitigating domain shift problem \cite{bib:DomainDet}. Raj~\emph{et al.} developed subspace alignment approach for detecting object in real-world scenarios \cite{bib:SubAlign}. For alleviating the problem of domain shift, Khodabandeh \emph{et al.} exploited a robust learning method with noisy labels \cite{bib:DomainAdapt}. Inoue \emph{et al.} proposed a cross-domain weakly-supervised training based on domain transfer and pseudo-labeling for domain adaptive object detection~\cite{bib:CrossWeakly}.

These works have indicated how to moderate the domain shift problem, but there has been relatively little work extensively studying the domain effect on detection performance. In contrast, based on underwater scenarios, we investigate the effect of quality-diverse data domain on object detection. Kalogeiton \emph{et al.} analyzed detection performance based on different image quality \cite{bib:DomainShift}, but we have advantages over their work: 1) \cite{bib:DomainShift} was reported before deep learning era, but we analyze deep learning-based object detection; 2) \cite{bib:DomainShift} considered the impact of simple factors (e.g., Gaussian blur), but our domain change is derived from realistic visual restoration; 3) \cite{bib:DomainShift} only analyzed cross-domain performance, but we investigate both cross-domain and within-domain performances; 4) our work contributes to aquatic robotics.

\section{Preliminary}
\subsection{Preliminary of Data Domain Based on Visual Restoration}

\begin{figure}[!t]
\centering
\includegraphics[width=8.cm]{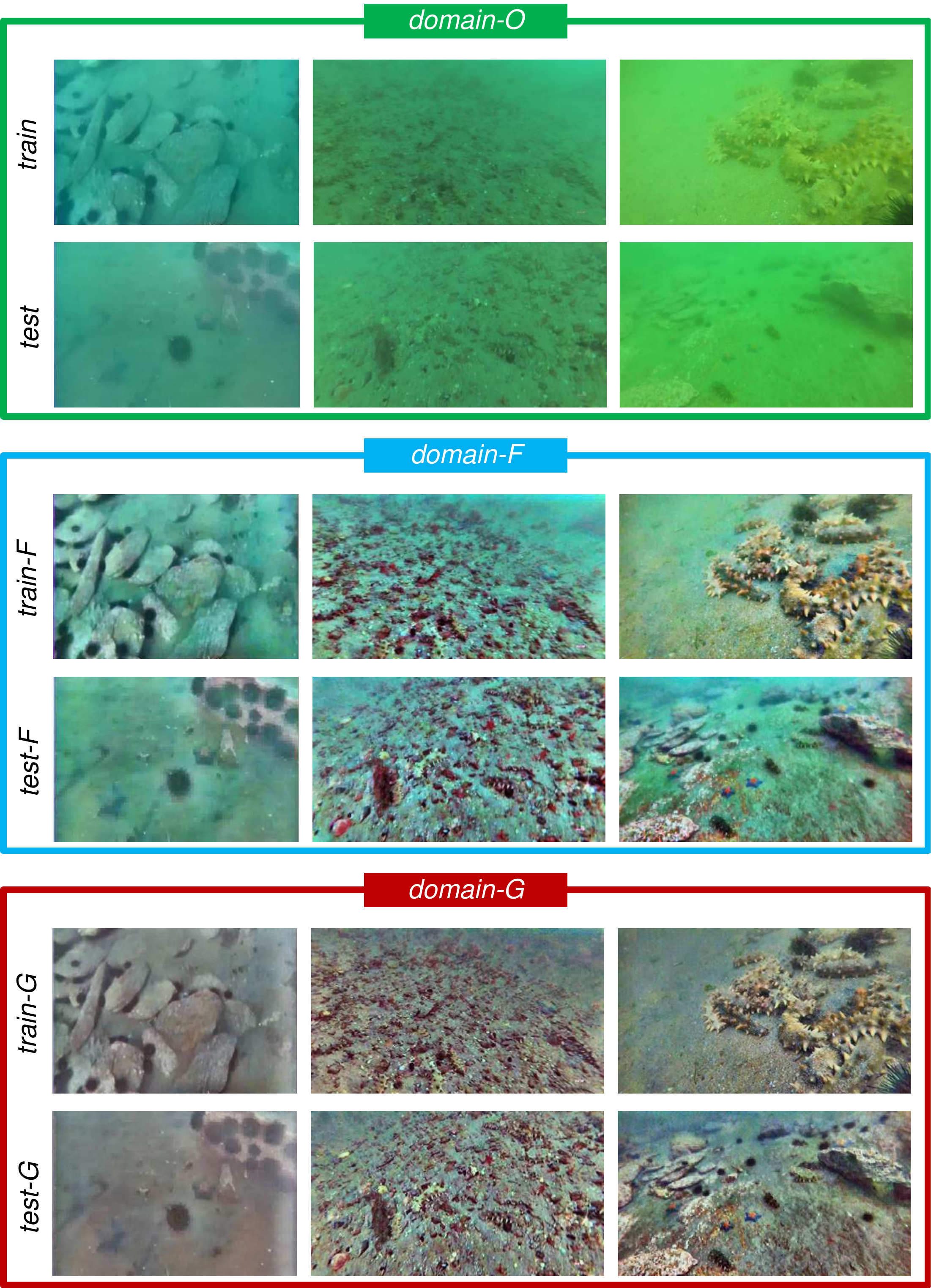}
\caption{Typical samples in \emph{domain-O}, \emph{domain-F}, and \emph{domain-G}.}
\label{fig:data}
\end{figure}

{\noindent\bf Domain generation.} The dataset is public available for underwater object detection, i.e., Underwater Robotic Picking Contest 2018\footnote{\url{http://en.cnurpc.org/}} (URPC2018). This dataset is collected on the natural seabed at Zhangzidao, Dalian, China. URPC2018 is composed of 2,901 aquatic images for training and 800 samples for testing. In addition, it contains four categories, i.e., ``trepang'', ``echinus'', ``shell'', and ``starfish''.

Based on URPC2018, three data domain are generated. 1) \emph{domain-O}: The original dataset with \emph{train} set and \emph{test} set; 2) \emph{domain-F}: All samples are processed by FRS, producing \emph{train-F} set for training and \emph{test-F} set for testing; 3) \emph{domain-G}: All samples are restored by GAN-RS, generating \emph{train-G} set for training and \emph{test-G} set for testing. Mixed \emph{train}, \emph{train-F}, and \emph{train-G} are denoted as \emph{train-all}. As shown in Fig.~\ref{fig:data}, \emph{domain-O} has strong color distortion, haziness, and low contrast. The degenerated visual samples are effectively restored in \emph{domain-F} and \emph{domain-G}.

\begin{figure}[!t]
\centering
\includegraphics[width=7.cm]{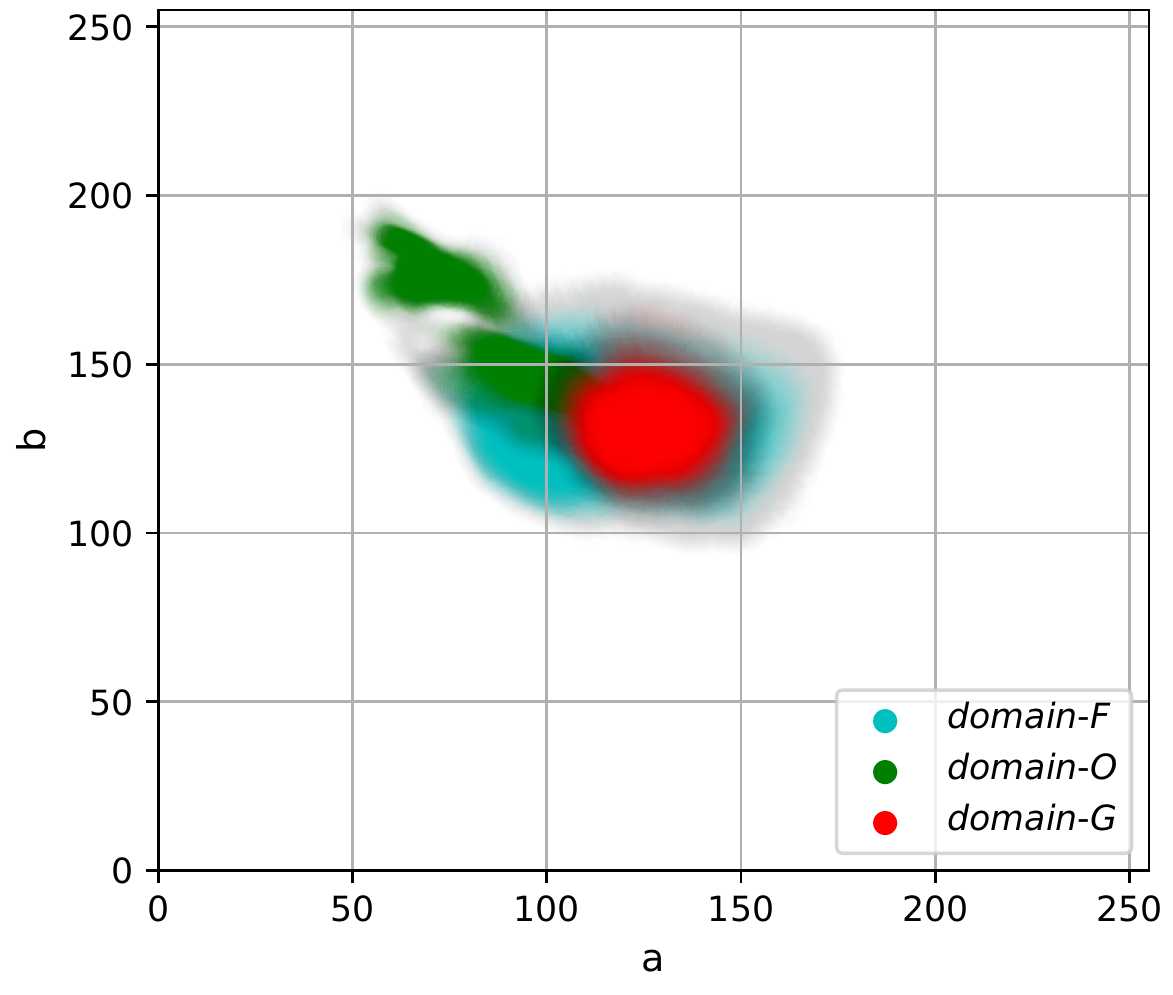}
\caption{Domain visualization in Lab color space. a-b distribution of \emph{domain-O} is concentrated and has color bias. In contrast, distributions of \emph{domain-F} and \emph{domain-G} are more scattered and have smaller biases. Color transparency indicates distribution probability.}
\label{fig:domain}
\end{figure}

\begin{table}[!t]
\renewcommand{\arraystretch}{1.2}
\setlength\tabcolsep{5.2pt}
\caption{Quality assessment for data domain.}
\label{tab:assessment}
\centering
\begin{tabular}{c | c c c c c}
\Xhline{1.5pt}
Domain          & UCIQE & UICM & UISM & UIConM & UIQM  \\
\hline
\emph{domain-O} & 0.39  & 0.20   & 3.86   & 0.12   & 1.58  \\
\emph{domain-F} &\bf0.56&\bf3.38 & 12.87  & 0.17   & 4.51  \\
\emph{domain-G} & 0.53  & 2.27   &\bf13.81&\bf0.18 &\bf4.78  \\
\Xhline{1.5pt}
\end{tabular}
\end{table}

{\noindent\bf Domain analysis.} According to \cite{bib:GAN-RS}, Lab color space has well ability to describe underwater properties of images. Thus, Fig.~\ref{fig:domain} illustrates a-b distribution in Lab color space. As a result, the distribution of \emph{domain-O} consistently gathers far from the color balance point (i.e., $(128,128)$). The bias between distribution center and the balance point means strong color distortion, and the concentrated distribution indicates strong haziness. On the contrary, different from \emph{domain-O}, the distributions of \emph{domain-F} and \emph{domain-G} have a trend of color balance and haze removal.

Underwater color image quality evaluation metric (UCIQE) \cite{bib:Ya15} and underwater image quality measure (UICM, UISM, UIConM, UIQM) \cite{bib:Pa15} are used to describe domain quality. UCIQE quantifies image quality via chrominance, saturation, and contrast. UIQM is a comprehensive quality representation of an underwater image, in which UICM, UISM, and UIConM separately describe color, sharpness, and contrast. Referring to Table~\ref{tab:assessment}, benefited from visual restoration, \emph{domain-F} brings about best UCIQE and UICM while \emph{domain-G} induces the best UISM, UIConM, and UIQM. Therefore, we define \emph{domain-F} and \emph{domain-G} as high-quality domains with high-quality samples. In contrast, \emph{domain-O} is defined as a low-quality domain with low-quality samples. Besides, referring to Fig.~\ref{fig:domain} and Table.~\ref{tab:assessment}, GAN-RS has better restoration results, so we define that GAN-RS induces a higher restoration intensity than FRS.

\subsection{Preliminary of Detector}
According to \cite{bib:RoIMix}, two-stage methods have no advantage over single-stage approaches on URPC2018. Therefore, because of the ability to induce both high accuracy and real-time inference speed, we leveraged single-stage detectors to perform underwater offline/online object detection. In detail, this paper investigates SSD, RetinaNet, RefineDet, and DRN. All these detectors are trained based on \emph{train}, \emph{train-F}, \emph{train-G}, or \emph{train-all}. As for training details, an SGD optimizer with $0.9$ momentum and $5\times 10^{-4}$ weight decay is employed, and batch size is $32$. We use the initial learning rate of $10^{-3}$ for the first $12\times 10^3$ iteration steps, then use the learning rate of $10^{-4}$ for the next $3\times 10^3$ steps and $10^{-5}$ for another $3\times 10^3$ steps. In this manner, all detectors can be sufficiently trained. For evaluation, mean average precision (mAP) is employed to describe detection accuracy.

\subsection{Preliminary of Aquatic Robot}
\begin{figure}[!t]
\centering
\includegraphics[width=8.cm]{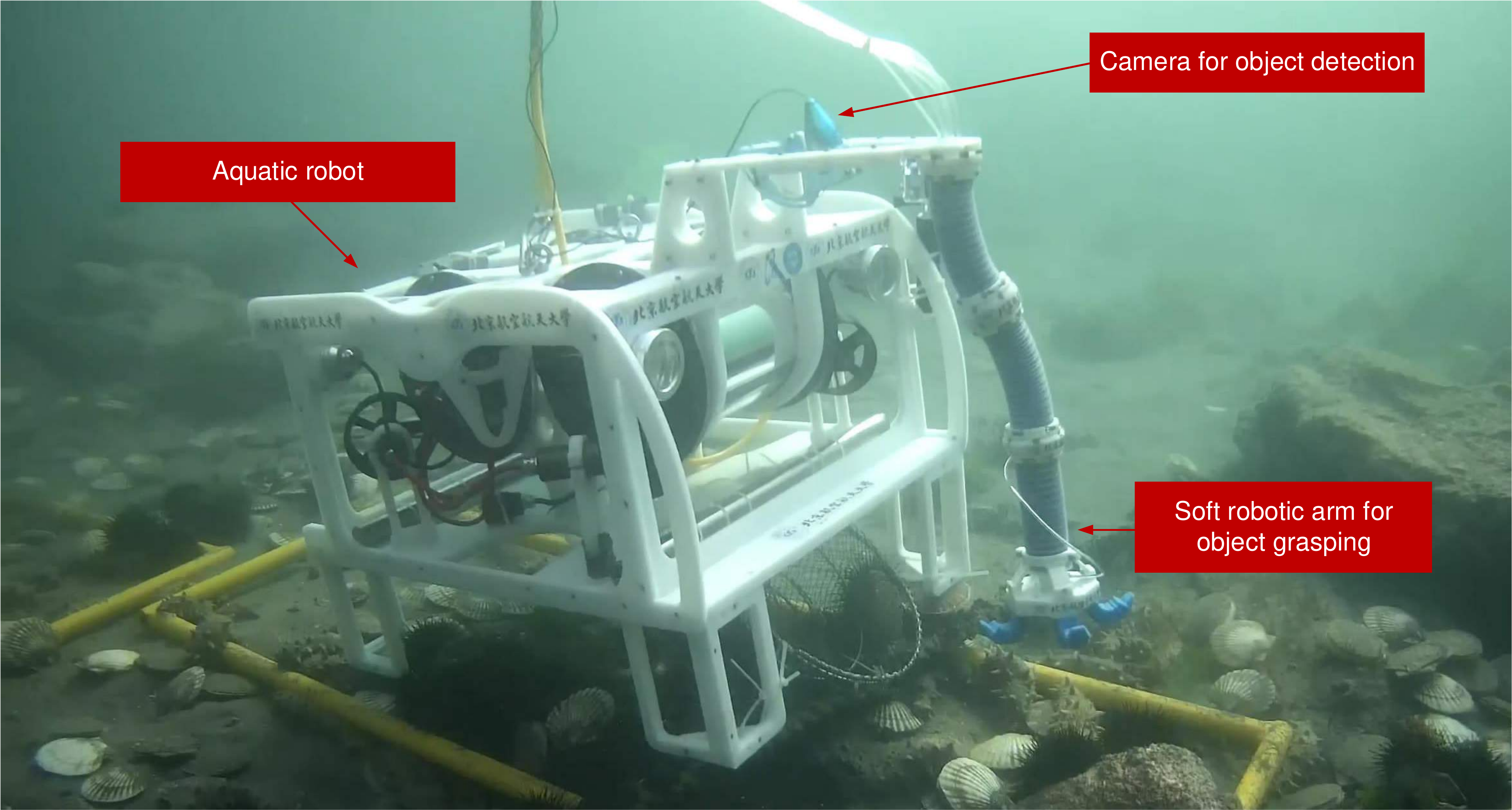}
\caption{Aquatic robot with visual perception system for object detection and grasping tasks.}
\label{fig:robot}
\end{figure}

As shown in Fig.~\ref{fig:robot}, the aquatic robot is equipped with a camera and a soft robotic arm for online object detection and grasping. It is $0.68$~m in length, $0.57$~m in width, $0.39$~m in height, and $50$~kg in weight. In the robot, we deploy a microcomputer with an Intel I5-6400 CPU, an NVIDIA GTX 1060 GPU, and 8 GB RAM as the processor. Thus, the robot has strong computing power for online object detection.

\begin{table}[!t]
\renewcommand{\arraystretch}{1.2}
\setlength\tabcolsep{1.8pt}
\caption{SSD detection results under conditions of different input sizes or backbones.}
\label{tab:SSD_det}
\centering
\begin{tabular}{c | c | c | c c c c c }
\Xhline{1.5pt}
method  & train data & test data & mAP & trepang & echinus & shell & starfish \\
\hline
\multirow{3}{2.3cm}{SSD320-VGG16} & \emph{train}    & \emph{test}    & \bf 69.3 & \bf 67.8 & \bf 84.9 & \bf 44.7 & \bf 79.7 \\
                                & \emph{train-F} & \emph{test-F}     & 67.8     & 68.9     & 82.3     & 42.2     & 78.0 \\
                                & \emph{train-G} & \emph{test-G}     & 65.9     & 65.4     & 82.3     & 39.0     & 76.9 \\
\hline
\multirow{3}{2.3cm}{SSD512-VGG16} & \emph{train}    & \emph{test}    & \bf 72.9 & \bf 70.2 & \bf 87.1 & \bf 50.8 & \bf 83.5 \\
                                & \emph{train-F} & \emph{test-F}     & 71.3     & 68.9     & 85.8     & 48.5     & 82.1 \\
                                & \emph{train-G} & \emph{test-G}     & 69.5     & 67.2     & 84.7     & 45.3     & 80.9 \\
\hline
\multirow{3}{2.3cm}{SSD512-MobileNet} & \emph{train}    & \emph{test}  & \bf 70.7 & \bf 65.3 & \bf 87.1 & \bf 47.5 & \bf 82.8 \\
                                    & \emph{train-F}  & \emph{test-F}  & 68.9     & 63.7     & 85.1     & 45.4     & 81.7 \\
                                    & \emph{train-G}  & \emph{test-G}  & 67.4     & 61.5     & 84.9     & 42.6     & 80.5 \\
\hline
\multirow{3}{2.3cm}{SSD512-ResNet101} & \emph{train}    & \emph{test}  & \bf 67.0 & 59.8     & \bf 86.3 & \bf 41.7 & \bf 80.3 \\
                                    & \emph{train-F}  & \emph{test-F}  & 65.6     & \bf 61.1 & 84.7     & 37.5     & 79.1 \\
                                    & \emph{train-G}  & \emph{test-G}  & 64.6     & 60.1     & 83.7     & 38.6     & 76.2 \\
\Xhline{1.5pt}
\end{tabular}
\end{table}

\section{Experimental Analysis}
\label{sec:EXP}
\subsection{Within-Domain Performance}
In this test, detectors' training and evaluation are based on identical data domain. The following analysis will unveil two points: 1) Domain quality has an ignorable effect on detection performance; 2) restoration is a thankless method for improving within-domain detection performance, because of the problem of low recall efficiency. Note that low recall efficiency means low precision under the condition of the same recall rate \cite{bib:SelectRefine}.

\begin{figure}[!t]
\centering
\subfigure { \label{fig:SSDVGG_a}
\includegraphics[width=8cm]{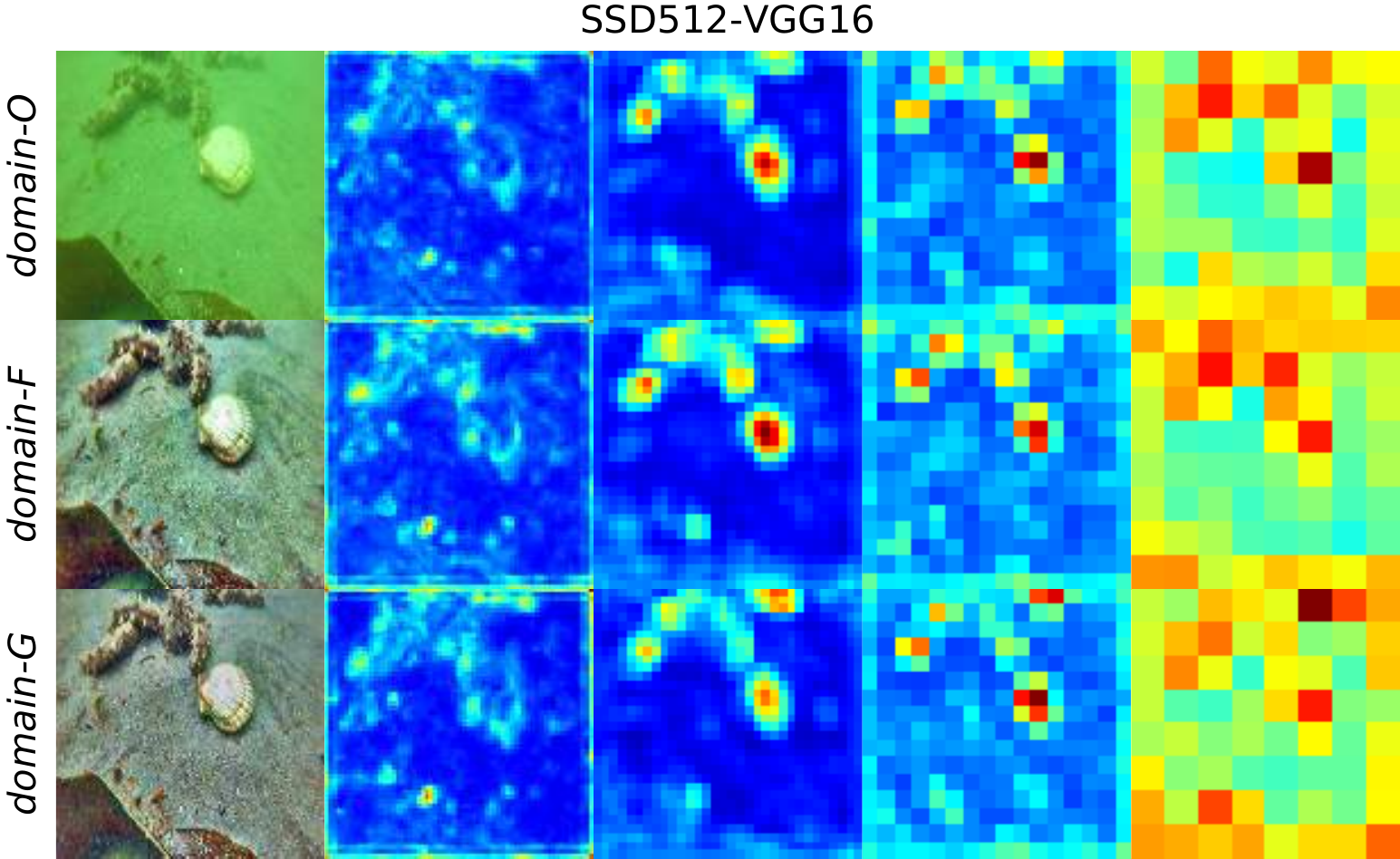}
}
\subfigure { \label{fig:DRNVGG_d}
\includegraphics[width=8cm]{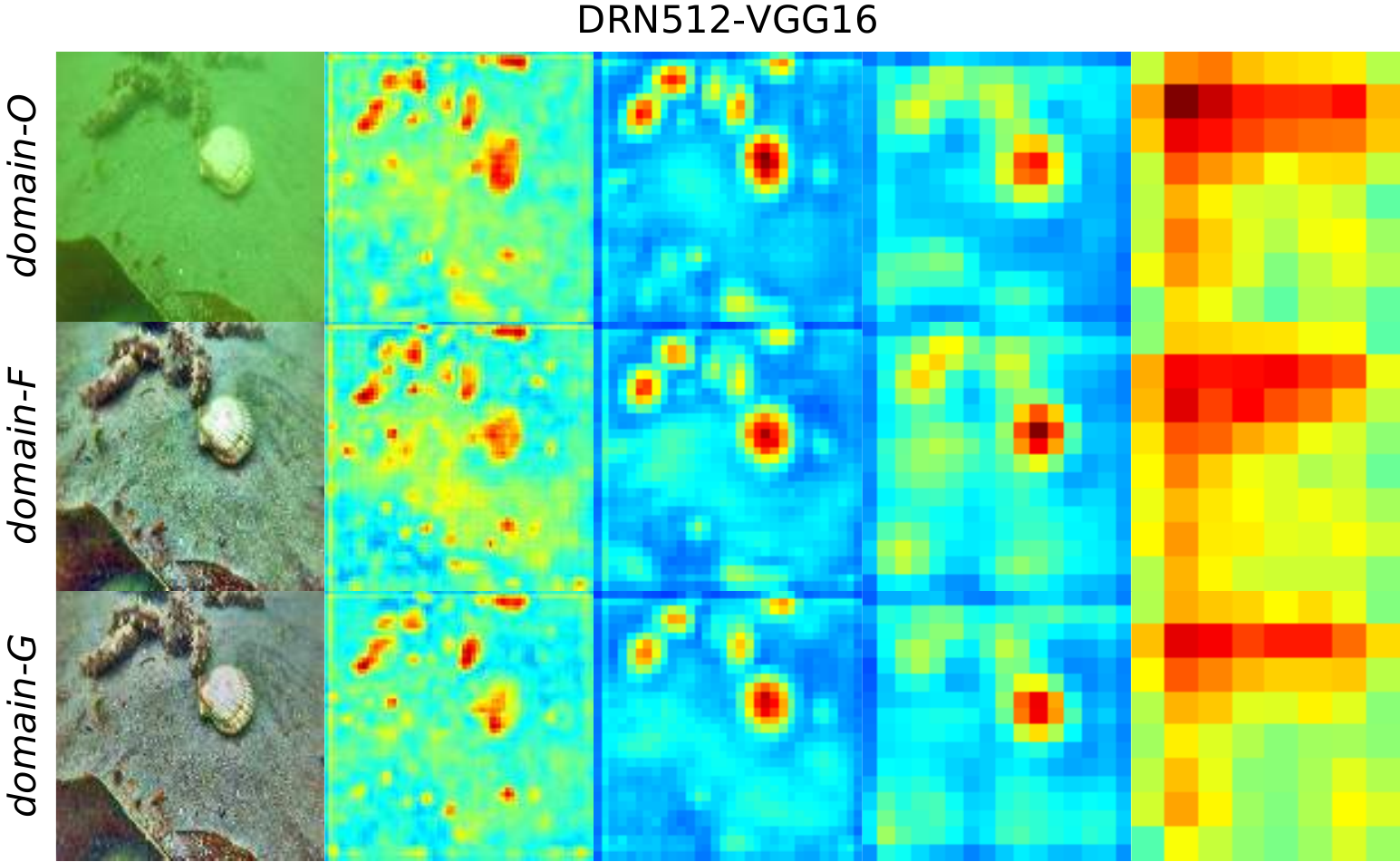}
}\caption{Visualization of convolutional representation for objects. each row contains input image and multi-scale features. High-level features are shown on the right. All features are precessed with L2 norm across channel, then they are normalized for visualization. For fair comparison, the same normalization factor is used for scale-identical features.}
\label{fig:featvis}
\end{figure}

\begin{table}[!t]
\renewcommand{\arraystretch}{1.2}
\setlength\tabcolsep{1pt}
\caption{Detection results based on RetinaNet, RefineDet, and DRN.}
\label{tab:other_det}
\centering
\begin{tabular}{c | c | c | c c c c c }
\Xhline{1.5pt}
method & train data & test data & mAP & trepang & echinus & shell & starfish \\
\hline
\multirow{3}{2.6cm}{RetinaNet512-VGG16}     & \emph{train}   & \emph{test}    & \bf 74.0   & \bf 69.8  & \bf 88.1 & \bf 54.7 & \bf 83.4 \\
                                            & \emph{train-F} & \emph{test-F}  & 72.5       & 69.1      & 87.1     & 50.7     & 82.9 \\
                                            & \emph{train-G} & \emph{test-G}  & 71.0       & 67.3      & 86.9     & 48.9     & 81.1 \\
\hline
\multirow{3}{2.6cm}{RefineDet512-VGG16}     & \emph{train}   & \emph{test}    & \bf 76.0   & \bf 73.8  & \bf 90.2 & \bf 54.1 & \bf 85.8 \\
                                            & \emph{train-F} & \emph{test-F}  & 72.9       & 72.0      & 88.6     & 46.4     & 84.6 \\
                                            & \emph{train-G} & \emph{test-G}  & 72.0       & 71.4      & 88.4     & 46.3     & 81.8 \\
\hline
\multirow{3}{2.6cm}{DRN512-VGG16}           & \emph{train}   & \emph{test}    & \bf77.1   & \bf 75.6  & \bf 91.1 & \bf 55.1 & \bf 86.7 \\
                                            & \emph{train-F} & \emph{test-F}  & 75.4      & 73.6      & 89.8     & 52.7     & 85.6 \\
                                            & \emph{train-G} & \emph{test-G}  & 73.8      & 72.0      & 89.8     & 49.9     & 83.5 \\
\Xhline{1.5pt}
\end{tabular}
\end{table}

\begin{figure*}[!t]
\begin{center}
\includegraphics[width=16cm]{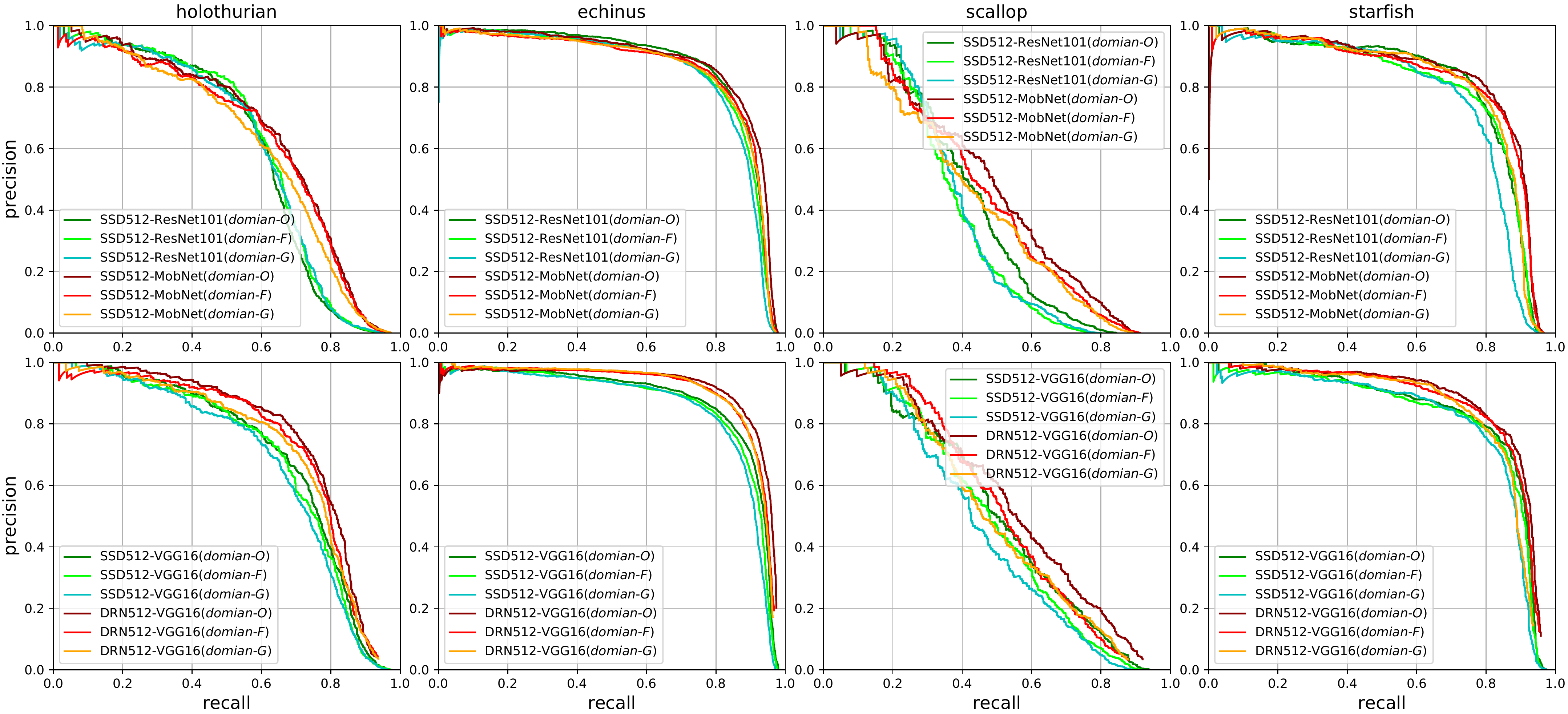}   
\caption{Precision-recall curves. For high precision (e.g., $>0.9$), domain difference has an ignorable effect on detection performance. Overall, \emph{domian-F} and \emph{domian-G} reduce recall efficiency so that lower average precision is induced.}
\label{fig:PR}
\end{center}
\end{figure*}

{\noindent\bf Numerical analysis.} At first, we train and evaluate SSD with different input sizes (i.e., 320 and 512) and backbones (i.e., VGG16 \cite{bib:VGG}, MobileNet \cite{bib:MobileNet}, and ResNet101 \cite{bib:ResNet}). As shown in Table~\ref{tab:SSD_det}, on \emph{domain-O}, \emph{domain-F}, and \emph{domain-G}, SSD320-VGG16 achieves mAP of $69.3\%, 67.8\%, 65.9\%$, and SSD512-VGG16 obtains mAP of $72.9\%, 71.3\%, 69.5\%$. It is seen that the accuracy decreases with the rise of restoration intensity. From backbone-variable assessments, the same trend emerges. Note that ResNet101 performs inferiorly to VGG16 and MobileNet, because large receptive field in ResNet101 is unfavorable to an immense number of small objects in URPC2018. Referring to Table~\ref{tab:other_det}, all of RetinaNet512, RefineDet512, and DRN512 can achieve the highest mAP on \emph{domain-O} and see the lowest mAP on \emph{domain-G}. Thus, in terms of mAP, detection accuracy is negatively correlated with domain quality. However, mAP cannot reflect accuracy details, so the following analysis will continue investigating within-domain performance.

{\noindent\bf Visualization of convolutional representation.} The human perceives domain quality based on object saliency. As a result, compared to low-quality domain, the human can more easily detect objects in high-quality domain since high-quality samples contain salient object representation. Thereby, we are inspired to investigate object saliency in CNN-based detectors. Fig.~\ref{fig:featvis} demonstrates multi-scale features in SSD and DRN. These features serve as the input of detection heads, so they are final convolutional features for detection. Referring to Fig.~\ref{fig:featvis}, despite of domain diversity, there is relatively little difference in object saliency in multi-scale feature maps. Hence, in terms of object saliency, domain quality has an ignorable effect on convolutional representation.

\begin{table}[!t]
\renewcommand{\arraystretch}{1.2}
\setlength\tabcolsep{1.1pt}
\caption{Cross-domain evaluation. $\downarrow$ is with respect to within-domain performance of the same test set.}
\label{tab:cde}
\centering
\begin{tabular}{c | c | c | c c c c c c }
\Xhline{1.5pt}
method &  train data & test data & mAP & trepang & echinus & shell & starfish \\
\hline
\multirow{4}{2.3cm}{SSD512-VGG16} & \multirow{2}{0.9cm}{\emph{train}} & \multirow{2}{0.7cm}{\emph{test-G}} & 52.1 & 42.5 & 70.2 & 36.6 & 59.0 \\
&   &   & $\downarrow$ 17.4  &$\downarrow$ 24.7 & $\downarrow$ 14.5 &$\downarrow$ 8.7 & $\downarrow$ 21.9 \\
\cline{2-8}
                                  & \multirow{2}{0.9cm}{\emph{train-G}}& \multirow{2}{0.7cm}{\emph{test}}  & 23.5 & 15.5 & 42.3 & 12.9 & 23.3 \\
&   &   & $\downarrow$ 49.4  &$\downarrow$ 54.7 & $\downarrow$ 44.8 &$\downarrow$ 37.9 & $\downarrow$ 60.2 \\
\hline
\hline
\multirow{4}{2.3cm}{DRN512-VGG16}     & \multirow{2}{0.9cm}{\emph{train}}   & \multirow{2}{0.7cm}{\emph{test-G}} & 57.9 & 53.7 & 74.2 & 40.0 & 63.7 \\
&   &   & $\downarrow$ 15.9  &$\downarrow$ 18.3 & $\downarrow$ 15.6 &$\downarrow$ 9.9 & $\downarrow$ 19.8 \\
\cline{2-8}
                                      & \multirow{2}{0.9cm}{\emph{train-G}} & \multirow{2}{0.7cm}{\emph{test}}   & 20.8 & 7.5  & 44.5 & 13.6 & 17.3 \\
&   &   & $\downarrow$ 56.3 &$\downarrow$ 68.1 & $\downarrow$ 46.6 &$\downarrow$ 41.5 & $\downarrow$ 69.4 \\
\cline{2-8}
\Xhline{1.5pt}
\end{tabular}
\end{table}

\begin{table}[!t]
\renewcommand{\arraystretch}{1.2}
\setlength\tabcolsep{1.5pt}
\caption{Cross-domain training. $\downarrow$ and $\uparrow$ are with respect to within-domain performance of the same test set.}
\label{tab:cdt}
\centering
\begin{tabular}{c | c | c | c c c c c c }
\Xhline{1.5pt}
method & train data & test data & mAP & trepang & echinus & shell & starfish \\
\hline
\multirow{6}{2.cm}{SSD512-VGG16}&\multirow{6}{1cm}{\emph{train-all}}& \multirow{2}{0.7cm}{\emph{test}}& 51.0      & 34.5    & 75.6     & 40.9    & 53.1 \\
&   &   & $\downarrow$ 21.9 &$\downarrow$ 35.7 & $\downarrow$ 11.5 &$\downarrow$ 9.9 & $\downarrow$ 30.4 \\
\cline{3-8}
                                   &                                & \multirow{2}{0.7cm}{\emph{test-F}}          & 71.4 & 69.2 & 85.4 & 48.4 & 82.4 \\
&   &   & $\uparrow$ 0.1    &$\uparrow$ 0.3 & $\downarrow$ 0.4 &$\downarrow$ 0.1 & $\uparrow$ 0.3 \\
\cline{3-8}
                                   &                                & \multirow{2}{0.7cm}{\emph{test-G}}          & 67.3 & 63.8 & 83.0 & 45.5 & 76.9 \\
&   &   & $\downarrow$ 2.2  &$\downarrow$ 3.4 & $\downarrow$ 1.7 &$\uparrow$ 0.2 & $\downarrow$ 4.0 \\
\hline
\hline
\multirow{6}{2.cm}{DRN512-VGG16}&\multirow{6}{1cm}{\emph{train-all}}& \multirow{2}{0.7cm}{\emph{test}}   & 52.0 & 34.5 & 75.6 & 40.9 & 53.1 \\
&   &   & $\downarrow$ 25.1 &$\downarrow$ 41.1 & $\downarrow$ 15.5 &$\downarrow$ 14.2 & $\downarrow$ 33.6 \\
\cline{3-8}
                                   &                                & \multirow{2}{0.7cm}{\emph{test-F}} & 75.8 & 75.0 & 89.8 & 53.1 & 85.3 \\
&   &   & $\uparrow$ 0.4 &$\uparrow$ 1.4 & 0 &$\uparrow$ 0.4 & $\downarrow$ 0.3 \\
\cline{3-8}
                                   &                                & \multirow{2}{0.7cm}{\emph{test-G}} & 72.2 & 70.5 & 86.6 & 51.1 & 80.7 \\
&   &   & $\downarrow$ 1.6 &$\downarrow$ 1.5 & $\downarrow$ 3.2 &$\uparrow$ 1.2 & $\downarrow$ 2.8 \\
\Xhline{1.5pt}
\end{tabular}
\end{table}

\begin{figure*}[!t]
\centering
\includegraphics[width=16.cm]{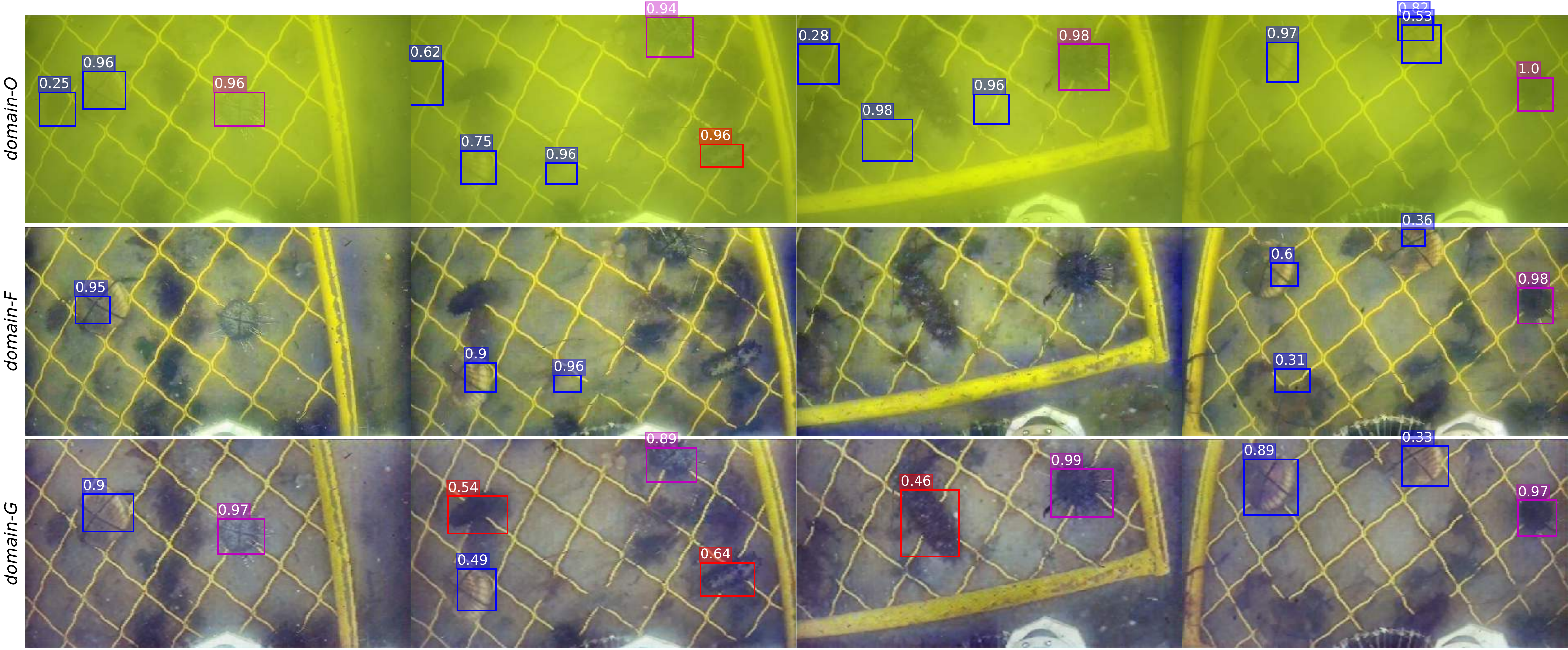}
\caption{Demonstration of online detection. DRN512-VGG16-$O$ and DRN512-VGG16-$F$ can hardly be qualified for this online detection. By suppressing the problem of domain shift, DRN512-VGG16-$G$ and GAN-RS perform better in this field underwater scene. Labels of the vertical axis denote training domains. ``trepang'', ``echinus'', and ``shell'' are detected in red, purple, and blue, respectively. Confidence scores are presented on the top-left of boxes.}
\label{fig:online_det}
\end{figure*}

{\noindent\bf Precision-recall analysis.} As shown in Fig.~\ref{fig:PR}, precision-recall curves are employed for further analysis of detection performance. It is can be seen that precision-recall curves have two typical appearances. On one hand, the high-precision part contains high-confident detection results, and here domain-related curves are highly overlapped. Referring to ``echinus'' detected by DRN512-VGG16, curves of \emph{domain-O}, \emph{domain-F}, and \emph{domain-G} cannot be separated when recall rate less than $0.6$. That is, when detecting high-confident objects, domain difference is negligible for detection accuracy. On the other hand, curves are separated in the low-precision part. In detail, the curve of \emph{domain-F} is usually below that of \emph{domain-O}, while the curve of \emph{domain-G} is usually below that of \emph{domain-F}. That is, when detecting hard objects (i.e., low-confident detection results), false positive increases with the rise of domain quality. For example, when recall rate equals $0.8$ in ``starfish'' detected by SSD512-VGG16, the precision of \emph{domain-F} is lower than that of \emph{domain-O}, and the precision of \emph{domain-G} is lower than that of \emph{domain-F}. Therefore, recall efficiency is gradually reduced with increasing restoration intensity.

Based on aforementioned analysis, it can be concluded that visual restoration impairs recall efficiency and is unfavorable for improving within-domain detection. In addition, because domain-related mAP is relatively close and high-confident recall is far more important than low-confident recall in robotic perception, we conclude that domain quality has an ignorable effect on within-domain object detection.

\subsection{Cross-Domain Performance}
In this test, detectors are trained and evaluated on different data domains. The following analysis will expose three viewpoints: 1) It is widely accepted that domain shift induces significant accuracy drop; 2) For cross-domain inference, learning based on low-quality domain has better generalization ability towards high-quality domain; 3) in domain-mixed learning, low-quality domain has smaller contribution so that low-quality samples cannot be well learned.

{\noindent\bf Cross-domain evaluation.} We use \emph{domain-O} and \emph{domain-G} for evaluation of direction-related domain shift. That is, we train detectors on \emph{train} and evaluate them on \emph{test-G}, or vice versa. As shown in Table~\ref{tab:cde}, mAP of all categories seriously declines. As a result, if \emph{train} and \emph{test-G} are employed, SSD512-VGG16 suffers $17.4\%$ mAP drop while DRN512-VGG16 encounters $15.9\%$ decrease in mAP. However, if \emph{train-G} and \emph{test} are adopted, SSD and DRN would suffer from a more dramatic accuracy exacerbation, i.e., mAP drops of $49.4\%$ and $56.3\%$. According to different degrees of accuracy drop caused by direction-opposite domain shift, it is seen that the generalization of \emph{train} towards \emph{test-G} is better than that of \emph{train-G} towards \emph{test}. Therefore, it can be concluded that compared to high-quality domain, low-quality domain induces better cross-domain generalization ability.

{\noindent\bf Cross-domain training.} For exploring detection performance with domain-mixed learning, we use \emph{train-all} to train detectors then evaluate them on \emph{test}, \emph{test-F}, and \emph{test-G}. Referring to Table~\ref{tab:cdt}, on \emph{test-F} and \emph{test-G}, SSD512-VGG16 and DRN512-VGG16 perform on-par with their within-domain performances. However, both SSD512-VGG16 and DRN512-VGG16 see dramatically worse accuracies on \emph{test}, i.e., $>20\%$ mAP drop. With the same training settings, within-domain performances can be similarly produced on high-quality \emph{domain-F} and \emph{domain-G}, but low-quality \emph{domain-O} suffers from significant accuracy decline. That is, when \emph{train-all} is adopted, samples in \emph{train} lose their effects to some extent. Thus, we conclude that cross-domain training is thankless for improving detection performance. Moreover, quality-diverse data domain has different contributions to the learning process so that low-quality samples cannot be well learned if mixed with high-quality samples.

\subsection{Domain-Effect in Robotics}
In this test, we conduct real-world experiments with the aquatic robot. The test venue is the natural seabed, located at Jinshitan, Dalian, China. The following analysis will answer the question $-$ \emph{how does visual restoration contribute to object detection?}

{\noindent\bf Online object detection in aquatic scenes.} Based on our aquatic robot, we use DRN512-VGG16 to detect underwater objects. According to different training domain, we denote detection methods as DRN512-VGG16-\emph{O}, DRN512-VGG16-\emph{F}, and DRN512-VGG16-\emph{G}, which are trained on \emph{train}, \emph{train-F}, \emph{train-G}, respectively. If DRN512-VGG16-\emph{F} or DRN512-VGG16-\emph{G} is employed, corresponding visual restoration (i.e., FRS or GAN-RS) should also be adopted to cope with online data. As shown in Fig.~\ref{fig:online_det}, DRN512-VGG16-\emph{O} almost completely loses its effect on object perception. Besides, DRN512-VGG16-\emph{F} and FRS also have difficulty in detecting underwater objects. In contrast, DRN512-VGG16-\emph{G} and GAN-RS have higher recall rate and detection precision in this real-world task. Because of the same detection method and content of training data, the huge performance gap should be caused by training domain. The experimental video is available at \url{https://youtu.be/RekqnNa0JY0}.

\begin{figure}[!t]
\centering
\includegraphics[width=8.cm]{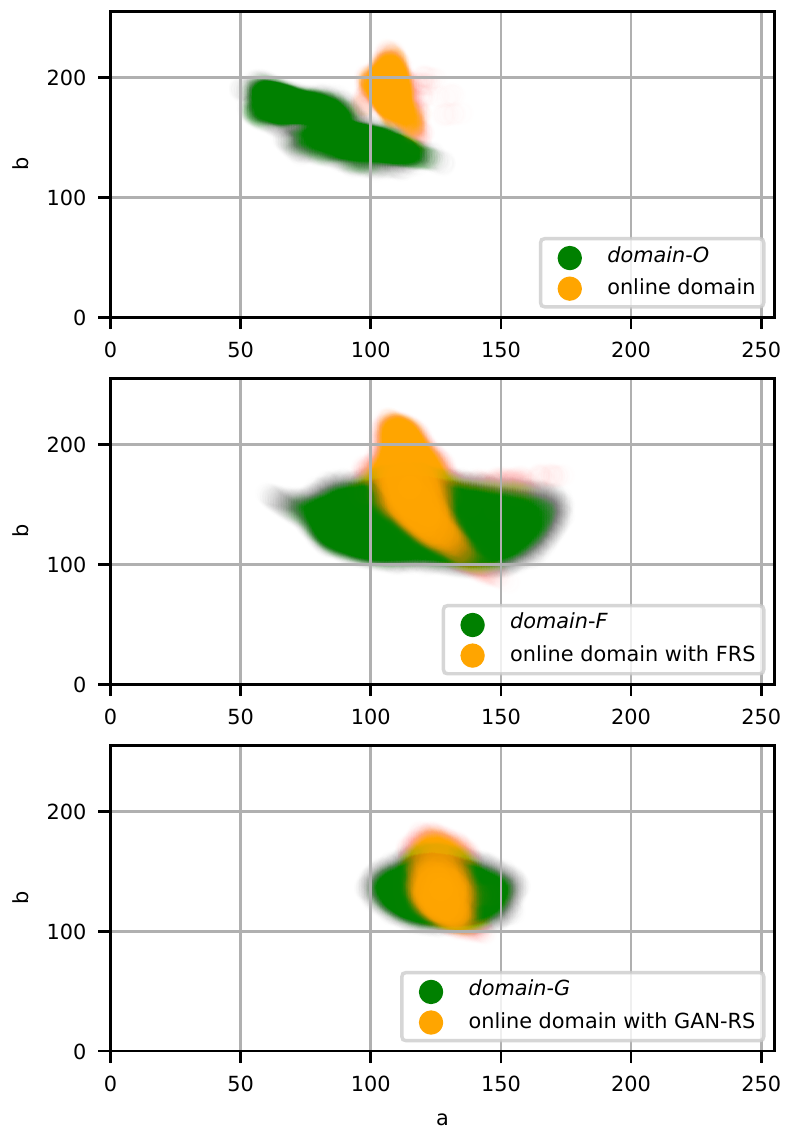}
\caption{Comparison of online domain and training domains in Lab color space. Color transparency indicates distribution probability.}
\label{fig:oline_domain}
\end{figure}

{\noindent\bf Online domain analysis.} As shown in Fig.~\ref{fig:oline_domain}, there is a huge discrepancy between online domain and \emph{domain-O}. Thus, DRN512-VGG16-\emph{O} suffers from serious degeneration on detection accuracy. Domain shift is moderated by FRS, but FRS is not sufficient to preserve detection performance in this scenario. On the contrary, GAN-RS has higher restoration intensity. As a result, processed by GAN-RS, online domain and \emph{domain-G} are highly overlapped as illustrated in Fig.~\ref{fig:oline_domain}. Therefore, DRN512-VGG16-\emph{G} and GAN-RS are able to perform this detection task well. It can be seen that the problem of domain shift is gradually solved with increasing restoration intensity. In addition, underwater scene domains are manifold (see Fig.~\ref{fig:int}), so domain-diverse data collection is unattainable. Therefore, contributing to domain shift suppression, visual restoration is essential for object detection in underwater environments.

\subsection{Discussion}
This paper has exposed phenomena of domain-related detection learning, and we discuss the following points to inspire future works.

{\noindent\bf Recall efficiency.} In within-domain tests, high-quality domain induces lower detection performance, because of low recall efficiency. Thus, high-quality domain incurs more false positives. However, object candidates that could bring about false positives exist in both training and testing phase. Under this condition, it is seen that the learning of these candidates is insufficient. Therefore, we advocate further research on how these candidates separately impact training and inference for exploring more efficient learning methods.

{\noindent\bf CNN's domain selectivity.} In cross-domain training, low-quality samples lose their effects so that accuracy drops on \emph{test} set. It is seen that the learning of CNN is characterized by domain selectivity. That is, samples' contributions are different in CNN-based detection learning. Therefore, we advocate further research on CNN's domain selectivity for building more robust real-world detectors.

\section{Conclusion}
\label{sec:CON}
In this paper, we have taken aim at domain analysis based on visual restoration and object detection for underwater robotic perception. Firstly, quality-diverse data domains are derived from URPC2018 dataset with FRS and GAN-RS. Furthermore, single-shot detectors are trained and evaluated, where within-domain and cross-domain performance are unveiled. Finally, we conduct online object detection to reveal the effect of visual restoration on object detection. As a result, we conclude novel viewpoints as follows: 1) Domain quality has an ignorable effect on within-domain convolutional representation and detection accuracy; 2) low-quality domain induces high cross-domain generalization ability; 3) low-quality domain can hardly be well learned in a domain-mixed learning process; 4) visual restoration is a thankless method for elevating within-domain performance, and it incurs relatively low recall efficiency; 5) visual restoration is essential in online robotic perception since it can relieve the problem domain shift.

In the future, we will further explore domain-related recall efficiency and learning selectivity. Additionally, more robotic tasks will be carried out based on our analysis.


\begin{thebibliography}{99}
\bibitem{bib:SoftArm}
Z.~Gong, J.~Cheng, X.~Chen, W.~Sun, X.~Fang, K.~Hu, Z.~Xie, T.~Wang, and L.~Wen, ``A bio-inspired soft robotic arm: Kinematic modeling and hydrodynamic experiments,'' \emph{J. Bionic Eng.}, vol. 15, no. 2, pp. 204--219, 2018.

\bibitem{bib:Manipulator}
M.~Cai, Y.~Wang, S.~Wang, R.~Wang, Y.~Ren, and M.~Tan, ``Grasping marine products with hybrid-driven underwater vehicle-manipulator system,'' \emph{IEEE Trans. Autom. Sci. Eng.}, doi: 10.1109/TASE.2019.2957782.

\bibitem{bib:ImageServoing}
J.~Gao, A.~A.~Proctor, Y.~Shi, and C.~Bradley, ``Hierarchical model predictive image-based visual servoing of underwater vehicles with adaptive neural network dynamic control,'' \emph{IEEE Trans. Cybern.}, vol. 46, no. 10, pp. 2323--2334, 2015.

\bibitem{bib:VisualSLAM}
A.~Kim and R.~M.~Eustice, ``Real-time visual SLAM for autonomous underwater hull inspection using visual saliency,'' \emph{IEEE Trans. Robot.}, vol. 29, no. 3, pp. 719--733, 2013.

\bibitem{bib:RoboFish}
Y.~Hu, W.~Zhao, and L.~Wang, ``Vision-based target tracking and collision avoidance for two autonomous robotic fish,'' \emph{IEEE Trans. Ind. Electron.}, vol. 56, no. 5, pp. 1401--1410, 2009.

\bibitem{bib:Sc04}
Y.-Y.~Schechner and N.~Karpel, ``Clear underwater vision,'' in \emph{Proc. IEEE  Conf. Comput. Vis. Pattern Recognition}, Washington, USA, Jun. 2004, pages I-536--I-543.

\bibitem{bib:UnderwaterBench}
R.~Liu, X.~Fan, M.~Zhu, M.~Hou, and Z.~Luo, ``Real-world underwater enhancement: Challenges benchmarks and solutions,'' \emph{arXiv:1901.05320}, 2019.


\bibitem{bib:Li16}
C.~Li, J.~Guo, R.~Cong, Y.~Pang, and B.~Wang, ``Underwater image enhancement by dehazing with minimum information loss and histogram distribution prior,'' \emph{IEEE Trans. Image Process.}, vol.~25, no.~12, pp. 5664--5677, 2016.

\bibitem{bib:Pe17}
Y.-T.~Peng and P.~C.~Cosman, ``Underwater image restoration based on image blurriness and light absorption,'' \emph{IEEE Trans. Image Process.}, vol.~26, no.~4, pp. 1579--1594, 2017.

\bibitem{bib:Ch17}
X.~Chen, Z.~Wu, J.~Yu, and L.~Wen, ``A real-time and unsupervised advancement scheme for underwater machine vision,'' in \emph{Proc. IEEE Int. Conf. Cyber Technol. Autom., Control, Intell. Syst.}, Hawaii, USA, Aug. 2017, pp. 271--276.






\bibitem{bib:GAN-RS}
X.~Chen, J.~Yu, S.~Kong, Z.~Wu, X.~Fang, and L.~Wen, ``Towards Real-Time Advancement of Underwater Visual Quality with GAN,'' \emph{IEEE Trans. Ind. Electron.}, vol. 66, no. 12, pp. 9350--9359, 2019.

\bibitem{bib:SSD}
W.~Liu, D.~Anguelov, D.~Erhan, C.~Szegedy, S.~Reed, C.~Y.~Fu, and A.~C.~Berg, ``SSD: Single shot multibox detector,'' in \emph{Proc. Eur. Conf. Comput. Vis.}, Amsterdam, Netherlands, Oct. 2016, pp. 21--37.

\bibitem{bib:RefineDet}
S.~Zhang, L.~Wen, X.~Bian, Z.~Lei, and S.~Z.~Li, ``Single-shot refinement neural network for object detection,'' in \emph{Proc. IEEE Conf. Comput. Vis. Pattern Recognition}, Salt Lack City, USA, Jun. 2018, pp. 4203--4212.

\bibitem{bib:RetinaNet}
T.~Y.~Lin, P.~Goyal, R.~Girshick, K.~He, and P.~Dollar, ``Focal loss for dense object detection,'' in \emph{Proc. IEEE Int. Conf. Comput. Vis.}, Venice, Italy, Oct. 2017, pp. 2980--2988.

\bibitem{bib:DRN}
X.~Chen, X.~Yang, S.~Kong Z.~Wu, and J.~Yu, ``Dual refinement network for single-shot object detection,'' in \emph{Proc. Int. Conf Robot. Autom.}, Montreal, Canada, May 2019, pp. 8305--8310.

\bibitem{bib:TDR}
L.~Pang, Z.~Cao, J.~Yu, P.~Guan, X.~Rong, H.~Chai, ``A visual leader-following approach with a TDR framework for quadruped robots,'' \emph{IEEE Trans. on Syst., Man, and Cybern. Syst.}, doi: 10.1109/TSMC.2019.2912715, 2019.

\bibitem{bib:SIFT}
D.-G.~Lowe, ``Distinctive image features from scale-invariant keypoints,'' \emph{Int. J. Comput. Vis.}, vol.~60, no.~2, pp. 91--110, 2004.



\bibitem{bib:SelectRefine}
C.~Chi, S.~Zhang, J.~Xing, Z.~Lei, S.~Z.~Li, and X.~Zou, ``Selective refinement network for high performance face detection,'' in \emph{Proc. AAAI Conf. Artifical Intell.}, Honolulu, USA, Jul. 2019, pp. 8231--8238.

\bibitem{bib:ACoupleNet}
Y.~Zhu, C.~Zhao, H.~Guo, J.~Wang, X.~Zhao, and H.~Lu, ``Attention couplenet: Fully convolutional attention coupling network for object detection,'' \emph{IEEE Trans. Image Process.,} vol. 28, no. 1, pp. 113--126, 2019.

\bibitem{bib:ExtremeNet}
X.~Zhou, J.~Zhuo, and P.~Kr{\"a}henb{\"u}hl, ``Bottom-up object detection by grouping extreme and center points,'' in \emph{Proc. IEEE Conf. Comput. Vis. Pattern Recognition}, Long Beach, USA, Jun. 2019, pp. 850--859.

\bibitem{bib:DomainRCNN}
Y.~Chen, W.~Li, C.~Sakaridis, D.~Dai, L.~Van Gool, ``Domain adaptive faster R-CNN for object detection in the wild,'' in \emph{Proc. IEEE Conf. Comput. Vis. Pattern Recognition}, Salt Lack City, USA, Jun. 2018, pp. 3339--3348.

\bibitem{bib:DomainDet}
J.~Xu, S.~Ramos, D.~V\'azquez, and A.~M.~L\'opez, ``Domain adaptation of deformable part-based models,'' \emph{IEEE Trans. Pattern Anal. Mach. Intell.}, vol. 36, no. 12, pp. 2367--2380, 2014.

\bibitem{bib:SubAlign}
A.~Raj, V.~P.~Namboodiri, and T.~Tuytelaars, ``Subspace alignment based domain adaptation for RCNN detector,'' \emph{arXiv:1507.05578}, 2015.

\bibitem{bib:DomainAdapt}
M.~Khodabandeh, A.~Vahdat, M.~Ranjbar, and W.~G.~Macready, ``A robust learning approach to domain adaptive object detection,'' \emph{Proc. IEEE Int. Conf. Comput. Vis.}, Seoul, Korea, Oct. 2019, pp. 480--490.

\bibitem{bib:CrossWeakly}
N.~Inoue, R.~Furuta, T.~Yamasaki, and K.~Aizawa, ``Cross-domain weakly-supervised object detection through progressive domain adaptation,'' in \emph{Proc. IEEE Conf. Comput. Vis. Pattern Recognition}, Salt Lack City, USA, Jun. 2018, pp. 5001--5009.

\bibitem{bib:DomainShift}
V.~Kalogeiton, V.~Ferrari, and C.~Schmid, ``Analysing domain shift factors between videos and images for object detection,'' \emph{IEEE Trans. Pattern Anal. Mach. Intell.}, vol. 38, no. 11, pp. 2327--2334, 2016.

\bibitem{bib:pix2pix}
P.~Isola, J.-Y.~Zhu, T.~Zhou, and A.-A.~Efros, ``Image-to-image translation with conditional adversarial networks,'' in \emph{Proc. IEEE Conf. Comput. Vis. Pattern Recognition}, Hawaii, USA, Jul. 2017, pp. 1125--1134.

\bibitem{bib:CycleGAN}
J.-Y.~Zhu, T.~Park, P.~Isola, and A.-A.~Efros, ``Unpaired image-to-image translation using cycle-consistent adversarial networks,'' in \emph{Proc. Int. Conf. Comput. Vis.}, Venice, Italy, Oct. 2017, pp. 2223--2232.

\bibitem{bib:RoIMix}
W.~H.~Lin, J.~X.~Zhong, S.~Liu, T.~Li, and  G.~Li, ``RoIMix: Proposal-fusion among multiple images for underwater object detection,'' \emph{arXiv:1911.03029}, 2019.

\bibitem{bib:Ya15}
M.~Yang and A.~Sowmya, ``An underwater color image quality evaluation metric,'' \emph{IEEE Trans. Image Process.}, vol.~24, no.~12, pp. 6062--6071, 2015.

\bibitem{bib:Pa15}
K.~Panetta, C.~Gao, and S.~Agaian, ``Human-visual-system-inspired underwater image quality measures,'' \emph{IEEE J. Ocean. Eng.}, vol.~41, no.~3, pp. 541--51, 2015.

\bibitem{bib:VGG}
K.~Simonyan and A.~Zisserman, ``Very deep convolutional networks for large-scale image recognition,'' \emph{arXiv:1409.1556}, 2014.

\bibitem{bib:ResNet}
K.~He, X.~Zhang, S.~Ren, and J.~Sun, ``Deep residual learning for image recognition,'' in \emph{Proc. IEEE Conf. Comput. Vis. Pattern Recognition}, Las Vegas, USA, Jun. 2016, pp. 770--778.

\bibitem{bib:MobileNet}
A.~Howard, M.~Zhu, B.~Chen, D.~Kalenichenko, W.~Wang, T.~Weyand, M.~Andreetto, and H.~Adam, ``Mobilenets: Efficient convolutional neural networks for mobile vision applications,'' \emph{arXiv:1704.04861,} 2017.

\end{thebibliography}
\end{document}